\documentclass{article}
\usepackage{ijcai19}

\usepackage{times}
\usepackage{soul}
\usepackage{url}
\usepackage[hidelinks]{hyperref}

\usepackage[utf8]{inputenc}
\usepackage[small]{caption}
\usepackage{graphicx}
\usepackage{amsmath}
\usepackage{booktabs}
\usepackage{algorithm}
\usepackage{algorithmic}
\graphicspath{{./Figures/}}
\usepackage{multicol}
\usepackage{multirow}
\usepackage{fancyhdr}

\usepackage{amsthm}
\usepackage{amsfonts,amssymb}

\title{Non-iterative Label Propagation in Optimal Leading Forest \footnote{This work has been supported by the National Key Research and Development Program of China under grants XXXX and XXXX, the National Natural Science Foundation of China under grant XXXX.}}

\author{
Ji Xu$^1$
\and
Guoyin Wang$^2$
\affiliations
$^1$Institute of Machine Intelligence, Guizhou University of Engineering Science, Bijie 551700, China\\
$^2$Chongqing Key Laboratory of Computational Intelligence, Chongqing University of Posts and Telecommunications,
Chongqing 400065, China\\
\emails
alanxuch@hotmail.com,
wanggy@ieee.org
}

\begin{document}

\maketitle

\begin{abstract}
 Graph based semi-supervised learning (GSSL) has intuitive representation and can be improved by exploiting the matrix calculation. However, it has to perform iterative optimization to achieve a preset objective, which usually leads to heavy computing burden. Another inconvenience lying in GSSL is that when new data come, the learning procedure has to be conducted from scratch. We leverage the partial order relation induced by the local density and distance between the data, and develop a highly efficient non-iterative label propagation algorithm based on a novel data structure named as optimal leading forest. The major two weaknesses of the traditional GSSL are addressed by this study. Experiments on various datasets have shown the promising efficiency and accuracy of the proposed method.
\end{abstract}

\section{Introduction}

Labels of data are laborious or expensive to obtain, while unlabeled data are generated or sampled in tremendous size in this big data era. This is the reason why semi-supervised learning (SSL) is constantly drawing the interests and attention from the machine learning society \cite{zhang2018semi,chen2018tri}. Among the variety of many SSL model streams, Graph-based SSL (GSSL) has the reputation of being easily understood through visual representation and is convenient to improve the learning performance by exploiting the corresponding matrix calculation. Therefore, there have been a lot of research works in this regard, e.g., \cite{liu2010large}, \cite{ni2012learning}, \cite{wang2017learning}.

However, the existing GSSL models have two apparent limitations. One is the models usually need to solve an optimization problem in an iterative fashion, hence the low efficiency. The other is that these models have difficulty in delivering the labels for a new bunch of data, because the solution for the unlabeled data is derived specially for the given graph. With newly included data, the graph has changed and the whole iterative optimization process is required to run from scratch.

We ponder the possible reasons of these limitations and argue that the crux is that these models take the relationship among the neighboring data points as ``peer-to-peer". Because the data points are considered equally significant to represent their class, most GSSL objective functions try optimizing on each data point with equal priority/weight. However, this ``peer-to-peer" relationship is questionable. For example, if a data point $x_c$ lies at the centering location of the space of its class, then it will has more representative power than the other one $x_d$ that diverges more from the central location, even if $x_c$ and $x_d$ are in the same $K$-NN or ($\epsilon$-NN) neighborhood. This idea is shared with many researchers. Recently, Li proposed a measure named as stability to ensemble clustering, in which they differentiated the objects within a cluster as core and halo \cite{Li2019Clustering}.

Since we have doubt in the ``peer-to-peer" relationship, this study is grounded on the partial-order-relation assumption: a)\emph{ the neighboring data points are not in equal status}, and b) \emph{the label of the leader (or parent) is the contribution of its followers (or children)}. Part a) of the assumption is explained above, we elaborate on Part b) a little here. The similar idea of ``a leader's label is the weighted summation of its followers' label" can be found in LLE\cite{Roweis2000Nonlinear} and AGR\cite{liu2010large}, and we will show in Section \ref{sec:LaPOLeaF} that Part b) of the assumption has solid mathematical foundation.

\begin{figure}[h]
  \centering
  \includegraphics[width=2.6in]{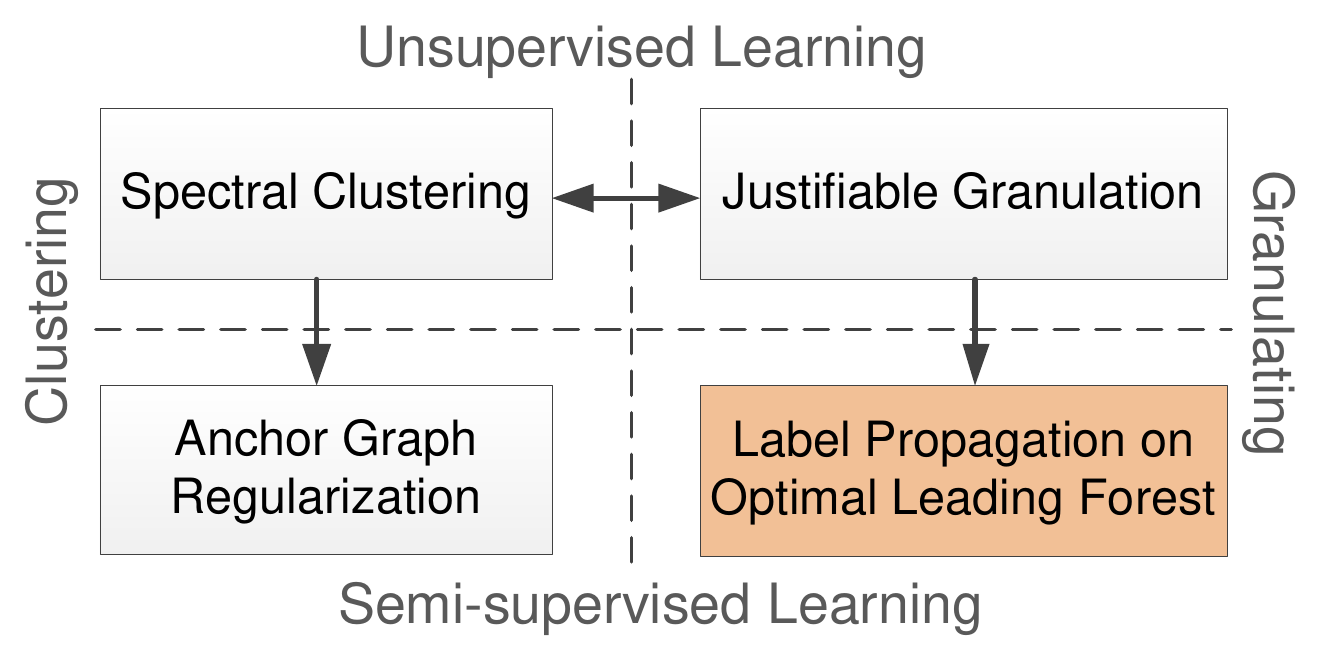}\\
  \caption{LoPOLeaF is derived from a concrete granulation method (LoDOG) that follows justifiable granulation principle, just as the GSSL method Anchor Graph Regularization derived from spectral clustering. }
  \label{fig:LocationOfTheWork}
\end{figure}

The mainstream methods of GSSL are closely related to spectral clustering \cite{Ng2001On}. Spectral clustering shares the same spirit with justifiable granulation principle \cite{Pedrycz2013Building} in granular computing (GrC) community. Among the several concrete granulation methods \cite{Pedrycz2015Data,Zhu2016Granular,SummitedtoTC}, local density based optimal granulation (LoDOG) \cite{SummitedtoTC} is characterized by its non-iterative fashion and high accuracy regardless the shapes of the information granules. Just as spectral clustering has inspired quite a few GSSL methods, LoDOG can be grounded on to develop a novel GSSL: Label Propagation on Optimal Leading Forest (LaPOLeaF). Fig. \ref{fig:LocationOfTheWork} shows the position where LaPOLeaF fits in the context formed by related existing works.

LaPOLeaF originates from LoDOG. In LoDOG, the input data was organized as an optimal number of subtrees and every non-center node in the subtrees is led by its parent to join the microcluster the parent belongs to. In \cite{Xu2016DenPEHC}, these subtrees are called \emph{Leading Tree}, so the collection of the optimal number of leading trees is called optimal leading forest (OLeaF). LaPOLeaF performs label propagation on the structure of the relatively independent subtrees in the forest, rather than on the traditional nearest neighbor graph. Therefore, LaPOLeaF exhibits several advantages when compared with other GSSL methods:

(a)	LaPOLeaF performs label propagation in a non-iterative fashion, so it is highly efficient.

(b) It is convenient to learn label for a newly arrived datum. 

(c) The leading relation between the samples reflects the evolution process from core to halo within a particular cluster, the interpretability of the learning result is therefore strengthened.

The rest of the paper is organized as follows. Section \ref{sec:RelatedWorks} briefly introduces some preliminaries. The model of LaPOLeaF is presented in details in Section \ref{sec:LaPOLeaF}. Section \ref{sec:ComplexAnalysis} analyzes the computation complexity, and Section \ref{sec:Experiments} describes the experimental study. We reach a conclusion in Section \ref{sec:Conclusion}.

\section{Preliminaries}\label{sec:RelatedWorks}

\subsection{Spectral Clustering and GSSL}


Spectral clustering \cite{Ng2001On,Luxburg2007A} maps the data points $\mathcal{X}=\{x_1,...,x_N\}$ and to the vertices $V=\{v_1,...,v_N\}$ of a graph $G$ and the similarity between two samples $x_i$ and $x_j$ to the weight $w_{ij}$ on the edge $e_{ij}=(v_i, v_j)$. When the weight $w_{ij}$ is below a certain threshold, then the edge may be cancelled. Therefore, an undirected graph $G=(V, E, W)$ can be constructed from a given dataset.

Although spectral clustering may have many variations, the core idea of them is common, which is to find way to partition the whole graph $G$ into $k$ sub-graphs, such that the summation of all the cuts between each subgraph and its complement is minimized. Formally, spectral clustering seeks to minimize
\begin{equation}
\label{eq:SpectralObj}
J({A_1},...,{A_k}) = \sum\limits_{i = 1}^k {\frac{{{\rm{cut}}({A_i},\overline {{A_i}} )}}{{|{A_i}|}}},
\end{equation}
where $A_i$ denotes the vertices in the $i^{th}$ subgraph, $ {\rm{cut}}({A_i},\overline {{A_i}} ) = \frac{1}{2}\sum\nolimits_{m \in {A_i},n \in \overline {{A_i}} } {{w_{mn}}} $.

Through introducing a specially designed matrix called Laplacian $L=D-W$, where $D$ is a diagonal matrix with elements $d_{ii}=\sum\limits_{j = 1}^k {{w_{ij}}} $, the cut minimization problem can be ingeniously transformed into a matrix eigenvalue decomposition problem and a succeeding plain k-means clustering for the derived eigenvectors.

Based on the same graph representation as in spectral clustering, GSSL propagates the label of the labeled samples $X_l$ to the unlabeled $X_u$, where $\mathcal{X}=X_l \cup X_u $. The propagation strength between $v_i$ and $v_j$ on each edge is in proportion to the weight $w_{i,j}$. Almost all the existing GSSL works on two assumptions: ``clustering assumption" and ``manifold assumption". Starting from the two assumptions, GSSL usually aims at optimizing an objective function with two terms. Liu proposed an Anchor Graph Regulation (AGR) approach to predict the label for each data point as a locally weighted average
of the labels of anchor points\cite{liu2010large}. To address the granularity dilemma in AGR , Wang proposed a hierarchical AGR method that adds a series of intermediate granular anchor layer between the finest original data and the coarsest anchor layer \cite{wang2017learning}. Recently, Du introduce the maximum correntropy criterion to make GSSL more robust to the datasets with noisy labels \cite{Du2018Robust}.

Slightly different from the two assumptions, Ni proposed a novel concept \emph{graph harmoniousness}, which integrates the feature learning and label learning into one framework called FLP \cite{ni2012learning}.

\subsection{Optimal Leading Forest}
Let $\boldsymbol{I}=\{1,2,...,N\}$ be the index set of $\mathcal {X}$, and $d_{ij}$ be the distance (under any metric) between $x_i$ and $x_j$.

\textbf{Definition 1} \cite{cite8}. The \textbf{local density} of $x_i$ is computed as
  ${\rho _i} = {\sum _{j \in {\bf{I}}\backslash \{ i\} }}{e^{ - {{(\frac{{{d_{ij}}}}{{{d_c}}})}^2}}}$, where $d_c$ is the cut-off distance or band-width parameter.

\textbf{Definition 2}. If $p_i$ is the nearest neighbor with higher local density to $x_i$, then $p_i$ is called the \textbf{leading node} of $x_i$. Formally, $l_i = \arg \mathop {\min }\limits_j \{ {d_{ij}}|{\rho _j} > {\rho _i}\}$, denoted as $x_{l_i}=\eta(x_i)$ for short. $d_{i,l_i}$ is called the \textbf{$\delta$-distance} of $x_i$, or simply $\delta_i$.

We store all the $x_{l_i}$ in an array named as \textbf{\emph{LN}}.

\textbf{Definition 3} \cite{Xu2016DenPEHC}. If $\rho_r=\mathop {\max }\limits_{1\le i \le N} \{\rho_i\}$, and $x_{l_i}=\eta(x_i)$, then one can draw an arrow starting from $x_i$, $i \in \boldsymbol{I}\backslash \{ r\}$ and ending at $x_{l_i}$ for $i \in \boldsymbol{I}$. Thus, $\mathcal{X}$ and the arrows form a tree $T$. Each node $x_i$ in $T$ (except $x_r$) tends to be led by $x_{l_i}$ to join the same cluster $x_{l_i}$ belongs to, unless $x_i$ itself makes a root (actually a center of the cluster represented by this subtree). Such a tree is called a \textbf{leading tree (LT)}.

\textbf{Definition 4} \textbf{($\boldsymbol \eta$ operator)} \cite{xu2017fat}. For any non-root node $x$ in an LT, there is a leading node $p$ for $x$. This mapping is denoted as $\eta(x)=p$.

we denote $\underbrace {\eta (\;\eta (\;...\eta (\bullet)))}_{n\;times} = {\eta ^n}(\bullet)$ for short.

\textbf{Definition 5} \textbf{(partial order in LT)} \cite{xu2017fat}. Suppose $x_i,x_j \in \mathcal{X}$, we say ${x_i} \prec {x_j}$, if and only if $\exists m \in {\mathbb{N}^+ }$  such that $x_j=\eta ^m(x_i)$.

\textbf{Definition 6} \textbf{(center potential)} \cite{cite8}. Let $\gamma_i$ be computed as $\gamma_i=\rho_i * \delta_i$, then $\gamma_i$ indicates the potential of $x_i$ to be selected as a center.

Intuitively, if an object $x_i$ has a large $\rho_i$ (means it has many near neighbors) and a large $\delta_i$ (means relatively far from another object of larger $\rho$), then $x_i$ would have great chance to be the center of a cluster.

From the view point of granular computing, clustering can serve as an approach to build information granules (IGs). There have been quite a few models to perform granulation, but how to evaluate the quality of the IGs remains an unaddressed issue until Pedrycz proposed \emph{the principle of justifiable granularity} \cite{Pedrycz2013Building,Pedrycz2015Data,Zhu2016Granular}. The principle indicates that a good information granule should has sufficient experiment evidence and specific semantic, which shares the same spirit with spectral clustering (see formula (\ref{eq:SpectralObj})).

Following this principle, LoDOG constructs the optimal IGs of $\mathcal{X}$ by disconnecting the corresponding leading tree into an optimal number of subtrees. The optimal number $N_g^{*}$ is derived via minimizing the objective function:

\begin{equation}\label{eq:ObjFunc}
  \mathop {\min }\limits_{N_g } \mathcal{Q}({N_g}) = \alpha  * H({N_g}) + (1 - \alpha )\sum\limits_{i = 1}^{{N_g}} {DCost({\Omega _i})},
\end{equation}

where $DCost({\Omega _i}) = \sum\limits_{j = 1}^{|{\Omega _i}| - 1} {\{ {\delta _j}|{\boldsymbol x _j} \in {\Omega _i}\backslash \{R(\Omega _i)\}\} }$.

Here, $N_g$ is the number of IGs; $\alpha$ is the parameter striking a balance between the experimental evidence and semantic; $\Omega _i$ is the set of points included in $i^{th}$ granule; $H(\bullet)$ is a strictly monotonically increasing function used to adjust the magnitude of $N_g$ to well match that of $\sum\limits_{i = 1}^{{N_g}} {DCost({\Omega _i})}$. This function can be selected from a group of common functions such as logarithm functions, linear functions, power functions, and exponential functions; $R({\Omega _i})$ is the root of the granule $\Omega _i$ as a leading tree.

We used LoDOG to construct the optimal leading forest (OLeaF) from the dataset. The readers are referred to \cite{SummitedtoTC} for more details of LoDOG.

\textbf{Definition 7}. $N_g^{*}$ leading trees can be constructed from the dataset $\mathcal{X}$ by using LoDOG method. All the leading trees are collectively called an \textbf{optimal leading forest (OLeaF)}.

The concept of OLeaF is used to localize the ranges of label propagation on the whole LT of $\mathcal{X}$. That is, OLeaF indicates where to stop propagating the label of a labeled datum to its partially ordered neighbors.

\section{Label Propagation on Optimal Leading Forest (LaPOLeaF)}\label{sec:LaPOLeaF}

LaPOLeaF first makes a global optimization to construct the OLeaF, then performs label propagation on each of the subtrees. With each step of the propagation, the label information is passed between the children and their parent, i.e., their common leading node.

Following the two assumptions of GSSL and taking the OLeaF structure into account , the objective of LaPOLeaF could be written as
\begin{equation}\label{Eq:LaPOLeaFObj}
\mathop {\min }\limits_L J(L) = \frac{1}{2}{\sum\limits_{i \in \boldsymbol{I} \backslash \{ R\} } {{W_i}\left\| {{L_i}{\rm{ - }}{L_{{p_i}}}} \right\|} ^2} + \mu \sum\limits_{i = 1}^l {{{\left\| {{L_i} - {Y_i}} \right\|}^2}},
\end{equation}
where $W_i$ is the similarity between $x_i$ and its leading node in the LT. LaPOLeaF only learns the label vector for unlabeled data, so the second term in the objective function can be removed. Thus (\ref{Eq:LaPOLeaFObj}) can be further simplified as:
\begin{equation}\label{Eq:LaPOLeaFObj2}
\mathop {\min }\limits_L J(L) = \frac{1}{2}{\sum\limits_{i \in \boldsymbol{I} \backslash \{ R\} } {{W_i}\left\| {{L_i}{\rm{ - }}{L_{{p_i}}}} \right\|} ^2}.
\end{equation}

\textbf{Theorem 1}. If consider $L_{p_i}$ as the only variables in (\ref{Eq:LaPOLeaFObj2}), then ${L_{{p_i}}} = \frac{{\sum\nolimits_i {{W_i}{L_i}} }}{{{W_i}}}$ is the optimal solution.

\begin{proof}
We have
\begin{displaymath}
\frac{{\partial J}}{{\partial {L_{{p_{_i}}}}}} = \frac{1}{2}\frac{{\partial \left( {\sum\nolimits_i {{W_i}{L_{{p_{_i}}}}^2}  - 2\sum\nolimits_i {{W_i}{L_{{p_{_i}}}}{F_i}} } \right)}}{{\partial {L_{{p_{_i}}}}}}
\end{displaymath}

\begin{equation}\label{Eq:OptiSoluProof1}
 = \sum\nolimits_i {{W_i}{L_{{p_{_i}}}} - } \sum\nolimits_i {{W_i}{L_i}}
\end{equation}
and
\begin{equation}\label{Eq:OptiSoluProof2}
\frac{{{\partial ^2}J}}{{\partial {L_{{p_{_i}}}}^2}} = {W_i} > 0,
\end{equation}
therefore,
\begin{equation}\label{Eq:OptiSoluProof3}
\frac{{\partial J}}{{\partial {L_{{p_{_i}}}}}} = 0 \Rightarrow {L_{{p_{_i}}}} = \frac{{\sum\nolimits_i {{W_i}{L_i}} }}{{\sum\nolimits_i {{W_i}} }}
\end{equation}
must be the optimal solution to the objective function (\ref{Eq:LaPOLeaFObj2}) and the proof is completed.
\end{proof}

 Following Theorem 1, the relationship between the children and their parent is formulated as (\ref{Eq:Assumption}), and the label propagation of LaPOLeaF will be guided mainly by this formula.

\begin{equation}\label{Eq:Assumption}
    {L_p} = \frac{{{{\sum\nolimits_i {{W_i} * L} }_i}}}{{\sum\nolimits_i {{W_i}} }},\;where\;{W_i} = \frac{{pop_i}}{{dist(i,p)}}.
\end{equation}

where $L_p$ is the label vector of the parent currently in consideration. $L_i$ is the label vector of the $i^{th}$ child w.r.t. the current parent. $pop_i$ is the population of the raw data points merged in the fat node $x_i$ in the subtree, if the node is derived as an information granule. If there is no granulation, all $pop_i$ are assigned with constant 1. We guarantee that $dist(i,p)>0$:  because $dist(i,p)=0$ implies that $x_i$ and $x_p$ are identical, then one can merge the two samples into $x_i$ and assign $pop_i=2$.

LaPOLeaF is designed to consist of three stages after the OLeaF has been constructed, namely, from children to parent (C2P), from root to root (R2R), and from parent to children (P2C).
%

To decide the layer index for each node, one can easily design a hierarchical traverse algorithm for the sub-leading-tree using the Queue data structure.

\subsection{Three Stages of Label Propagation in LaPOLeaF}
We introduce two pairs of definitions for discussing some properties.

 \textbf{Definition 8}. A node in the subtree of the OLeaF is an \textbf{unlabeled node} (or the node is unlabeled), if its label vector is \textbf{0}. Otherwise, i.e., if its label vector has any positive element, the node is called a \textbf{labeled node} (or the node is labeled).

 \textbf{Definition 9}. A subtree in OLeaF is called an \textbf{unlabeled subtree} (or the subtree is unlabeled), if every node in this tree is not labeled. Otherwise, this tree is called a \textbf{labeled subtree} (or the subtree is labeled).
\subsubsection{From Children to Parent}
The parent $p$ gets its label as the weighted summation of its children (Eq. \ref{Eq:Assumption}), and the label of the parent of $p$ is computed likely in a cascade fashion.

Since the label of a parent is regarded as the contribution of its children, the propagation process is required to start from the bottom of each subtree. The label vector of an unlabeled children is initialized as vector $\boldsymbol0$. Once the layer index of each node is ready, the bottom-up propagation can start to execute in a parallel fashion for the labeled subtrees.

\textbf{Theorem 2}. After C2P propagation, the root of a labeled subtree must be labeled.

\begin{proof}
A parent is labeled if it has at least one child labeled after the corresponding round of the propagation. The propagation is progressing sequentially along the bottom-up direction, and the root is the parent at the top layer. Therefore, this proposition obviously holds.
\end{proof}

\subsubsection{From Root to Root}\label{sec:R2RPropa}

If the labeled data are rare or unevenly distributed among the classes, there would be some unlabeled subtrees. In such a case, we must borrow some label information from other labeled subtrees. Because the label of the root is more stable than other nodes, the root of an unlabeled subtree $r_u$ should borrow label information from a root of a labeled subtree $r_l$. However, there must be some requirements for $r_l$. To keep consistence with our partial order assumption, $r_l$ is required to be superior to $r_u$ and is the nearest root to $r_u$. Formally,

\begin{equation}\label{Eq:r2rBorrow}
  {r_l(u)} = \mathop {arg\min }\limits_{{r_i} \in {R_L}} \{ dist({r_u},{r_i})|{r_u} \prec {r_i}\},
\end{equation}
where $R_L$ is the set of labeled roots.

If there exists no such $r_l$ for a particular $r_u$, we can conclude that the root $r_T$ of the whole leading tree constructed from $\mathcal{X}$ (before splitting into a forest) is not labeled. So, to guarantee every unlabeled root can successfully borrow a label, one needs to guarantee $r_T$ be labeled by assigning

\begin{equation}\label{Eq:rTBorrow}
    r_T^l = \mathop {arg\min }\limits_{{r_i} \in {R_L}} \{ dist({r_T},{r_i})\}
\end{equation}


\subsubsection{From Parent to Children}\label{sec:P2CPropa}
After the previous two stages, every root of the subtrees are labeled. In P2C propagation, the labels are propagated in a top-down fashion.

\emph{\textbf{Remark}: In the C2P propagation, the unlabeled node is labeled with zero vector, hence makes no contribution to the label of their leading node. However, in the P2C propagation, the label of an unlabeled node must be assumed to have some positive elements.}

There are two situations: \\a) for a parent $x_p$, all $m$ children $x_i$, $1 \le i \le m$, are unlabeled. Here, We simply assign $L_i$=$L_p$, because this assignment directly satisfies (\ref{Eq:Assumption}) no matter what value each $W_i$ takes. \\b) for a parent $x_p$, without loss of generality, assume the first $m_l$ children are labeled, and the other $m_u$ children are unlabeled ($m=m_l+m_u$). In this situation, we generate a virtual parent $x_{p'}$ to replace the original $x_p$ and the $m_l$ labeled children. Using (\ref{Eq:Assumption}), we have
\begin{equation}\label{Eq:VirtualNode1}
{L_p} = \frac{{\sum\nolimits_{i = 1}^{{m_l}} {{W_i}{L_i}}  + \sum\nolimits_{i = {m_l} + 1}^{{m_l} + {m_u}} {{W_i}{L_i}} }}{{\sum\nolimits_{i = 1}^m {{W_i}} }},
\end{equation}
Assuming all the $L_i$ of the unlabeled node are the same yields
\begin{equation}\label{Eq:VirtualNode3}
{L_p} - \frac{{\sum\nolimits_{i = 1}^{{m_l}} {{W_i}{L_i}} }}{{\sum\nolimits_{i = 1}^m {{W_i}} }} = \frac{{\sum\nolimits_{i = {m_l} + 1}^{{m_l} + {m_u}} {{W_i}} }}{{\sum\nolimits_{i = 1}^m {{W_i}} }}{L_i},
\end{equation}
Let
\begin{equation}\label{Eq:VirtualNode4}
{L_{p'}} = {L_p} - \frac{{\sum\nolimits_{i = 1}^{{m_l}} {{W_i}{L_i}} }}{{\sum\nolimits_{i = 1}^m {{W_i}} }},  \frac{1}{C} = \frac{{\sum\nolimits_{i = {m_l} + 1}^{{m_l} + {m_u}} {{W_i}} }}{{\sum\nolimits_{i = 1}^m {{W_i}} }}
\end{equation}

Then, we have ${L_i} = C{L_{p'}}$. Since the labeled vector is about which element is the greatest, the constant $C$ can be omitted.
Therefore, the $m_u$ unlabeled children can be assigned with the label $L_{p'}$ like in the first situation. That is,
\begin{equation}\label{Eq:P2CSolution}
{L_i} = L_{p'}={L_p} - \frac{{\sum\nolimits_{i = 1}^{{m_l}} {{W_i}{L_i}} }}{{\sum\nolimits_{i = 1}^m {{W_i}} }}.
\end{equation}

%

\subsection{LaPOLeaF Algorithm}
We present the overall algorithm of LaPOLeaF here, including some basic information about OLeaF construction.

\begin{algorithm}[tb]
\caption{LaPOLeaF Algorithm}
\label{alg:LaPOLeaF}
\textbf{Input}: Dataset $\mathcal{X}=\mathcal{X}_l \cup \mathcal{X}_u$\\
\textbf{Parameter}: $percent, \alpha, H(\bullet)$\\
\textbf{Output}: Labels for $\mathcal{X}_u$\\
 \hspace*{0.5cm} \textbf{Part 1}: \emph{//Preparing the OLeaF\;}
\begin{algorithmic}[1] 
\STATE Compute $Dist$ for $\mathcal{X}$.
\STATE Compute local density $\boldsymbol{\rho}$.
\STATE Compute leading nodes \textbf{\emph{LN}}, $\delta$-distance $\boldsymbol{\delta}$ .
\STATE Compute center potential $\boldsymbol \gamma$.
\STATE Split the LT into OLF using objective function (\ref{eq:ObjFunc}).
\STATE Build adjacent List for each subtree.\\
\textbf{Part 2}: \emph{//Label propagation on the OLeaF\;}
\STATE Decide the level index for each node.
\STATE C2P propagation using (\ref{Eq:Assumption}).
\STATE R2R propagation using (\ref{Eq:r2rBorrow}) and (\ref{Eq:rTBorrow}).
\STATE P2C propagation using (\ref{Eq:P2CSolution}).
\STATE \textbf{return} Labels for $\mathcal{X}_u$.
\end{algorithmic}
\end{algorithm}

\subsection{Deriving the Label for a New Datum}\label{sec:NewLabelX}
A salient advantage of LaPOLeaF is that it can obtain the label for a new datum (let us denote this task as LXNew) in $O(n)$ time. This is because: (a) the leading tree structure can be incrementally updated in $O(n)$ time, and the LoDOG algorithm can find $N_g^{*}$ in $O(n)$ time, OLeaF can be therefore updated in $O(n)$ time;  (b) the label propagation on the OLeaF takes $O(n)$ time.

The reader can refer to \cite{xu2017fat}, in which the authors provided an detailed description of the algorithm for incrementally updating the fat node leading tree and proved the correctness of the updating algorithm.

\section{Time Complexity Analysis }\label{sec:ComplexAnalysis}
By investigating each step in Algorithm \ref{alg:LaPOLeaF}, we find out that except the calculation of the distance matrix requires exactly $n(n-1)/2$ basic operations, all other steps in LaPOLeaF has the linear time complexity to the size of $\mathcal{X}$ . When compared to LLGC \cite{zhou2004learning}, FLP, AGR, HAGR, and RGSSL-MCC, LaPOLeaF is much more efficient, as listed in Table \ref{tab:ComplexityComp}. In Table \ref{tab:ComplexityComp}, $n$ is the size of $\mathcal{X}$; $T$ is the number of iterations; $K$ is the number of classes; $m$ is the number of anchors; $m_h$ is the number of points on the $h^{th}$ layer.

\begin{table}
  \centering
   \renewcommand{\arraystretch}{1.3}
  \caption{Complexity comparison}\label{tab:ComplexityComp}
  \begin{tabular}{|c|c|c|}
  \hline
  {\bf Methods} & {\bf Graph} & {\bf Label propagation} \\
   \hline
  LLGC  & $O(n^2)$ & $O(n^3)$  \\
   \hline
  FLP & $O(n^2)$ & $O(T_1Kn^2+T_2K^2n^2)$ \\
   \hline
  AGR  & $O(Tmn)$ & $O(m^2n+m^3)$ \\
   \hline
  HAGR  &$O(Tm_hn)$ & $O(m_h^2n+m_h^3)$ \\
   \hline
    RGSSL-MCC   &$O(n^2)$ & $O(Tn^3)$ \\
   \hline
  LaPOLeaF&$O(n^2)$ & $O(n)$ \\

  \hline
\end{tabular}
\end{table}

It is worthwhile to mention again that LaPOLeaF can obtain the label for a new datum in $O(n)$ time, while other GSSL methods cannot.

\section{Experimental Studies}\label{sec:Experiments}
The efficiency and effectiveness of LaPOLeaF have been evaluated on various datasets, we chose to report the results of 2 real world ones here due to page limitation. The information of the datasets is shown in Table \ref{tab:LaPOLeafDatasets}. The ImageNet2012\_s is used to demonstrate the efficiency and accuracy of LaPOLeaF. The two water quality datasets are used to show the capability of LaPOLeaF in time series prediction, due to its convenience in LXNew task.

 The experiments are conducted on a Dell P7920 work station with two Intel Xeon Silver 4110 CPUs, 16GB DDR4 memory, and an NVIDIA Quadro P2000 GPU.

\begin{table}
  \centering
    \renewcommand{\arraystretch}{1.3}
  \caption{Information of the datasets in the experiments}\label{tab:LaPOLeafDatasets}
  \begin{tabular}{|c|c|c|c|}
    \hline
   {\bf Dataset} & {\bf \# Instances} & {\bf \# Attributes} & {\bf \# Classes} \\
    \hline
     ImageNet\_s & 2481 & 4096 & 5 \\
    \hline
    \multicolumn{4}{c}{} \\[-4mm]
    \hline
    {\bf Dataset} & {\bf \# Instances} & {\bf \# Dimension} & {\bf task} \\
    \hline
    Water(HP) & 28,065 & \{5, 12\} & regression \\
    \hline
    Water(DO) & 28,065 & \{5, 12\} & regression \\
    \hline
  \end{tabular}
\end{table}


\begin{table}
  \centering
   \renewcommand{\arraystretch}{1.3}
  \caption{Parameters configuration for the 5 datasets}\label{tab:Para5Sets}
  \begin{tabular}{|c|c|c|c|}
    \hline
    {\bf Dataset} &  {\bf percent} & { $\boldsymbol \alpha$} & { $\boldsymbol {H(x)}$}  \\
    \hline
      ImageNet\_s & 3 & 0.4 & $80x\times1.001^x$  \\
    \hline
     Water & 5 & 0.5 & $0.1x$ \\
    \hline

  \end{tabular}
\end{table}

\subsection{Imagenet2012 Subsets}
Details of the dataset ImageNet\_s is listed in Table \ref{tab:ImageNetInfo}.
\begin{table}
  \centering
  \renewcommand{\arraystretch}{1.3}
  \caption{Detailed information of dataset ImageNet2012\_s }\label{tab:ImageNetInfo}
  \begin{tabular}{|c|c|c|}
    \hline
    {\bf File folder name} & {\bf Class label} & {\bf \# samples} \\
    \hline

    n01440764 &	tench &	454\\
     \hline
n01484850	&great white shark	&530\\
 \hline
n02096177&	cairn terrier&	480\\
 \hline
n03450230&	gown&	489\\
 \hline
n07932039&	eggnog&	528\\
  \hline

  \end{tabular}
  \end{table}

We first crop the images according to the box information carried by the XML files, then use the Caffe \cite{Jia2014Caffe} tools to convert each of the 2,481 images into a 4096-dimensional ``fc7" feature. Then, the dataset has been transformed into a $2481\times 4096$ matrix. The subsequent steps are of standard LaPOLeaF, with the parameter settings listed in Table \ref{tab:Para5Sets}. The learning process by LaPOLeaF is very fast, whose detailed time consumption information is listed in Table \ref{tab:TimeImageNetInfo}.

\begin{table}
  \centering
  \renewcommand{\arraystretch}{1.3}
  \caption{Time consumption (seconds) of LaPOLeaF stages for ImageNet2012\_s }\label{tab:TimeImageNetInfo}
  \begin{tabular}{|c|c|c|c|}
    \hline
    {\bf Preprocessing} & {\bf Distance} & {\bf OLeaF} & {\bf Propagation}\\
    \hline

    55 &	2.27	&1.73&	0.18\\
     \hline

  \end{tabular}
  \end{table}

We compare the accuracy on Imagenet2012 subsets with RGSSL-MCC when the data labels contain no noise and there are 10\%, 30\%, and 50\% labels respectively. Since LaPOLeaF is transductive, only the transductive results in RGSSL-MCC are compared. Table \ref{tab:ImageNetAcc} shows that LaPOLeaF achieved a higher accuracy on ImageNet\_s.

\begin{table}[!h]
 \centering
  \small
    \renewcommand{\arraystretch}{1.3}
  \caption{Accuracy comparison on dataset ImageNet2012\_s}\label{tab:ImageNetAcc}
  \begin{tabular}{|c|c|c|c|}
\hline
\multirow{2}{*}{\bf Method}&\multicolumn{3}{c|}{\bf Percentage of the labeled samples (\%)} \\
\cline{2-4}

&{10}&{ 30} &{ 50}  \\
\hline

 RGSSL-MCC &  $63\pm2.8$ &  $71.5\pm1.2$ & $74\pm0.8$ \\

\hline
LaPOLeaF & $\boldsymbol {92.6\pm0.8}$ &	$\boldsymbol {95.4\pm0.9}$&  $\boldsymbol {97.5\pm0.3}$ \\
\hline

  \end{tabular}
  \end{table}

\begin{figure}[h!]
  \centering
  \includegraphics[width=2.2in]{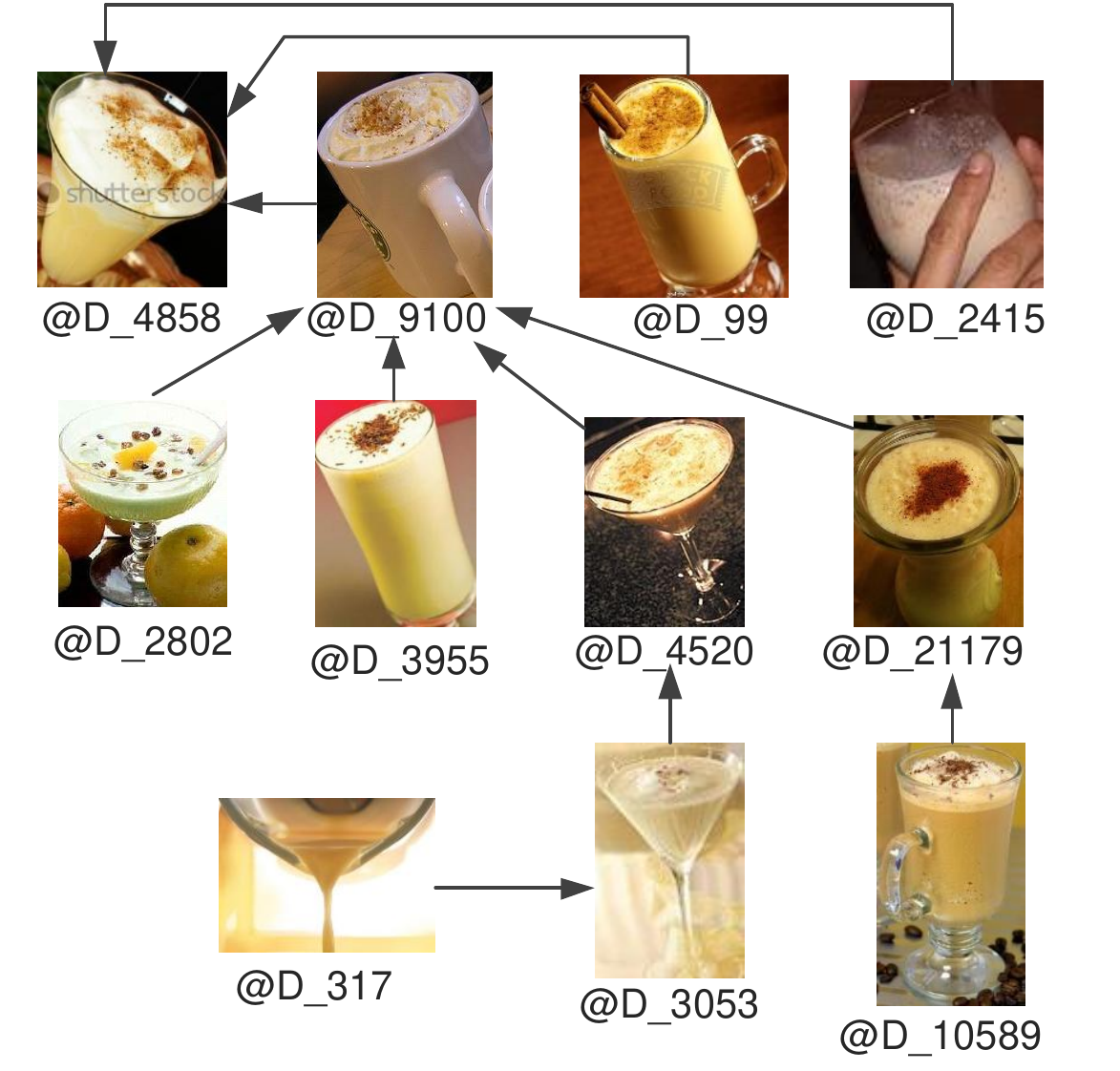}\\
  \caption{The inner structure found by LaPOLeaF reveals the evolution of the subtle difference among the same class. For the name of the images, @D=``n07932039" and the filename extension  ``.JPEG" is omitted. }
  \label{fig:Interpretabiliy}
\end{figure}

Apart from the accuracy and efficiency, LaPOLeaF discovered the subtle evolution within a given class (see Fig. \ref{fig:Interpretabiliy}). Thus, as a new GSSL, LaPOLeaF has good interpretability.

\subsection{Water Quality Prediction}
The water quality datasets are sampled in Chongqing, China, from March 3rd to November 21st in 2014. The sampling frequency is once per 15 minutes. We choose the PH and DO (dissolved oxygen) index to predict, and compare the performance of LaPOLeaF with least square support vector regression (LSSVR) \cite{Suykens2002Least} --- a classic model that has been widely applied in ecological environment \cite{Adnan2017Improving,Goyal2014Modeling}.

Water quality data are time series. We first fold both the PH and DO data as vectors of dimensionality 6 and 13, then regard the first 5 or 12 elements as observed attributes $x$ and the last element as response variable $y$. Thus, dataset Water(PH) is transformed into PH-5Attr and PH-12Attr, and similarly the dataset Water(DO). The first 10,000 records are taken as training set and the succeeding 1000 as test set. The parameter configuration used for LSSVR is $\{\gamma=0.5, \lambda=5, p=3\}$, while that for LaPOLeaF can be found in Table \ref{tab:Para5Sets}.

%

We will evaluate the performance of prediction via summation of squared error (SSE), which is defined as $SSE= \sum\nolimits_i {{{({y_i} - {{\widehat y}_i})}^2}}$, where ${y_i}$ is the ground truth and ${\widehat y}_i$ is the predicted value. Table \ref{tab:LoPOLeaFLSSVRComprare} shows the accuracy and efficiency of the two competing methods. The training of LaPOLeaF is faster than LSSVR. However, the prediction of LSSVR is faster than LaPOLeaF, because LSSVR directly computes the estimated value through a formula once the parameters are learned. By contrast, LaPOLeaF needs an $O(N)$ complexity to update the OLeaF with a new observed pattern. For example, when we perform prediction on the PH-12Attr dataset, LSSVR takes 2\emph{ms} and LaPOLeaF takes 5\emph{ms}. Although slower, LaPOLeaF could definitely meet the real-time-prediction requirement.

\begin{table}[!h]
 \centering
  \small
    \renewcommand{\arraystretch}{1.3}
  \caption{Prediction accuracy and training time of LaPOLeaF and LSSVR on the water quality data}\label{tab:LoPOLeaFLSSVRComprare}
  \begin{tabular}{|c|cccc|}
\hline
\multirow{2}{*}{\bf Method}&\multicolumn{4}{c|}{\bf Summation of squared error (SSE)} \\
\cline{2-5}

&{PH-5Attr}&{ PH-12Attr} &{ DO-5Attr} &{ DO-12Attr} \\
\hline

 LSSVR & {\bf 13.76}& {\bf 13.59} & 3168.9 &	5827.2 \\

\hline
LaPOLeaF & 21.70 &	21.06& {\bf	1166.6} & {\bf	1213.0}\\
\hline

\multicolumn{5}{c}{} \\[-4mm]
    \hline

    \multirow{2}{*}{\bf Method}&\multicolumn{4}{c|}{\bf Running time (s)} \\
\cline{2-5}

&{ PH-5Attr}&{ PH-12Attr} &{ DO-5Attr} &{ DO-12Attr} \\
\hline

 LSSVR & 5.67 &5.87 &	8.28	& 10.22 \\  	
\hline	 	 	
LaPOLeaF &{\bf 3.91 }	&{\bf 3.86} &{\bf 3.74}	&{\bf 3.77}\\
\hline
  \end{tabular}
  \end{table}

\begin{figure}[htbp]
  \centering
  \includegraphics[width=3.3in]{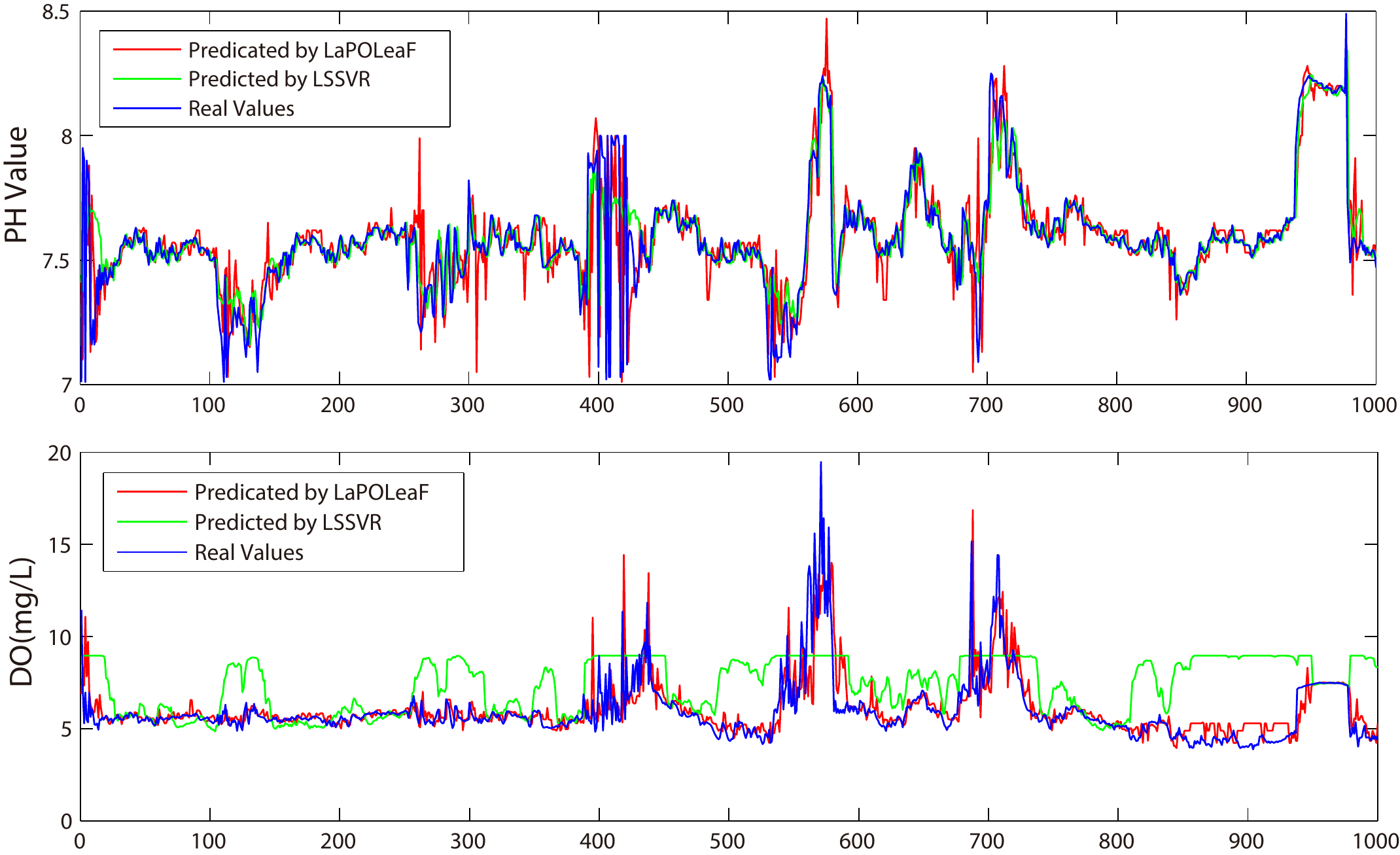}\\
  \caption{The accuracy comparison between LaPOLeaF and LSSVR for water quality (PH and DO values) prediction. }
  \label{fig:WaterPrediction}
\end{figure}

Also, the differences of the true value and the predicted valued are visualized in Fig. \ref{fig:WaterPrediction}, from which one can read that LaPOLeaF can approximate the ground truth value on both Water(PH) and Water(DO) datasets. On the dataset Water(DO), LaPOLeaF out performs LSSVR substantially. On Water(PH), LaPOLeaF achieved a slightly lower accuracy than LSSVR, yet the results are comparable. The possible reason for the different performances may be attributed to the spirit of the two methods.

\section{Conclusions}\label{sec:Conclusion}
This paper addressed the two weaknesses of the existing GSSL, namely, low efficiency caused by iterative optimization and inconvenience to predict the label for newly arrived data. We firstly made a sound assumption that the neighboring data points are not in equal positions, but lying in a partial order relation; and the label of a parent can be regarded as the contribution of its children. Based on this assumption and the granulation method named as LoDOG, a new non-iterative semi-supervised approach called LaPOLeaF is proposed. LaPOLeaF exhibits two salient advantages: a) It has much higher efficiency than the sate-of-the-art models while keep the promising accuracy. b) It can deliver the labels for a few newly arrived data in a time complexity of $O(N)$, where $N$ is the data size. When evaluated in classifying ImageNet2012 subset and predicting water quality, LaPOLeaF showed good accuracy and efficiency. The intermediate structure OLeaF helps the practitioner and user to better understand the learning result. We plan to extend LaPOLeaF in two directions: one is to scale it to accommodate big data, and the other is to improve its accuracy while keeping the high efficiency unchanged.

%
%

%
%
%

\bibliographystyle{named}
\bibliography{LaPOIGRefBaseIJCAI}

\end{document}